\documentclass[11pt,a4paper]{article}
\usepackage[hyperref]{acl2018}
\usepackage{times}
\usepackage{latexsym}
\usepackage{graphicx}
\usepackage{amsfonts}

\usepackage{url}
\aclfinalcopy % Uncomment this line for the final submission

\usepackage{fancyhdr}
\usepackage{geometry}
\usepackage{layout}
\newcommand{\ignore}[1]{}

\title{On the Practical Computational Power of Finite Precision RNNs \\ for Language Recognition}

\author{Gail Weiss \\
  Technion, Israel \\
  {\tt } \\\And
  Yoav Goldberg \\
  Bar-Ilan University, Israel \\
  \\
    {\tt \{sgailw,yahave\}@cs.technion.ac.il }\\
    {\tt yogo@cs.biu.ac.il} \\ 
    \\\And
  Eran Yahav \\
  Technion, Israel \\
    {\tt } \\}

\date{}

\begin{document}
\maketitle
\thispagestyle{fancy} %\maketitle triggers \thispagestyle{plain}, which suppresses the "accepted to ACL..." header

\begin{abstract}
While Recurrent Neural Networks (RNNs) are famously known to be Turing complete, this relies on infinite precision in the states and unbounded computation time. We consider the case of RNNs with finite precision whose computation time is linear in the input length. Under these limitations, we show that different RNN variants have different computational power. In particular, we show that the LSTM and the Elman-RNN with ReLU activation are strictly stronger than the RNN with a squashing activation and the GRU. This is achieved because LSTMs and ReLU-RNNs can easily implement counting behavior. We show empirically that the LSTM does indeed learn to effectively use the counting mechanism.
\end{abstract}

\section{Introduction}
\label{intro}

Recurrent Neural Network (RNNs) emerge as very strong learners of sequential data.
A famous result by Siegelmann and Sontag \shortcite{ss92,ss94}, and its extension in
\cite{siegelmann-book}, demonstrates that an Elman-RNN \cite{elman1990finding} with a sigmoid activation function, rational weights and infinite precision states can simulate a Turing-machine in real-time, making RNNs Turing-complete. Recently, Chen et al \shortcite{isi} extended the result to the ReLU activation function. 
However, these constructions (a) assume reading the entire input into the RNN state and only then performing the computation, using \emph{unbounded time}; and (b) rely on having \emph{infinite precision} in the network states. As argued by Chen et al \shortcite{isi}, this is not the model of RNN computation used in NLP applications.
Instead, RNNs are often used by feeding an input sequence into the RNN one item
at a time, each immediately returning a state-vector that corresponds to a
prefix of the sequence and which can be passed as input for a subsequent
feed-forward prediction network operating in constant time. The amount of tape
used by a Turing machine under this restriction is linear in the input length,
reducing its power to recognition of context-sensitive language. More
importantly, computation is often performed on GPUs with 32bit floating point
computation, and there is increasing evidence that competitive performance can
be achieved also for quantized networks with 4-bit weights or fixed-point
arithmetics \cite{binarized}. The construction of \cite{siegelmann-book} implements pushing 0 into a binary stack by the operation $g \leftarrow g/4 + 1/4$. This allows pushing roughly 15 zeros before reaching the limit of the 32bit floating point precision. Finally, RNN solutions that rely on carefully orchestrated  mathematical constructions are unlikely to be found using backpropagation-based training.

In this work we restrict ourselves to \emph{input-bound recurrent neural networks with finite-precision states (IBFP-RNN)}, trained using back-propagation.
This class of networks is likely to coincide with the networks one can expect to obtain when training RNNs for NLP applications. An IBFP Elman-RNN is finite state. But what about other RNN variants?
\begin{figure*}
\centering
\begin{tabular}{cc}
\includegraphics[width=0.42\textwidth]{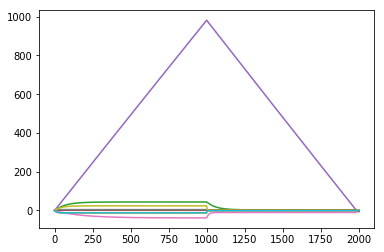} &
\includegraphics[width=0.42\textwidth]{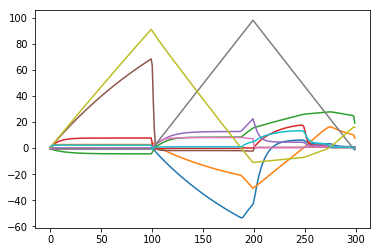}\\
(a) $a^nb^n$-LSTM on $a^{1000}b^{1000}$ &
(b) $a^nb^nc^n$-LSTM on $a^{100}b^{100}c^{100}$ \\
\hline
\includegraphics[width=0.42\textwidth]{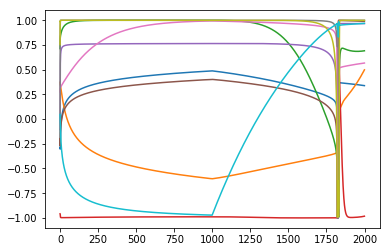} &
\includegraphics[width=0.42\textwidth]{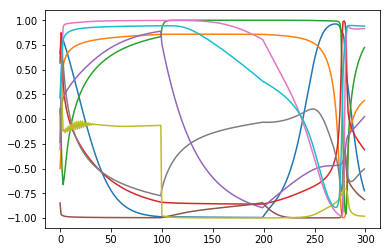} \\
(c) $a^nb^n$-GRU on $a^{1000}b^{1000}$ &
(d) $a^nb^nc^n$-GRU on $a^{100}b^{100}c^{100}$ \\
\end{tabular}
\caption{Activations --- c for LSTM and h for GRU --- for networks trained on $a^nb^n$ and $a^nb^nc^n$. The LSTM has clearly learned to use an explicit counting mechanism, in contrast with the GRU.}
\label{fig:plots}
\end{figure*}
In particular, we consider the Elman RNN (SRNN) \cite{elman1990finding} with
squashing and with ReLU activations, the Long Short-Term Memory (LSTM)
\cite{lstm} and the Gated Recurrent Unit (GRU)
\cite{cho2014learning,chung2014empirical}. 

The common wisdom is that 
the LSTM and GRU introduce additional \emph{gating components} that handle the vanishing gradients problem of training SRNNs, thus stabilizing training and making it more robust.
The LSTM and GRU are often considered as almost equivalent variants of each other.

We show that in the input-bound, finite-precision case, there is a real difference between the computational capacities of the LSTM and the GRU:
the LSTM can easily perform unbounded counting, while the GRU (and the SRNN) cannot. This makes the LSTM a variant of a k-counter machine \cite{fischer1968}, while the GRU remains finite-state. Interestingly, the SRNN with ReLU activation followed by an MLP classifier also has power similar to a k-counter machine.

These results suggest there is a class of formal languages that can be recognized by LSTMs but not by GRUs.
In section \ref{sec:exps}, we demonstrate that for at least two such languages, the LSTM manages to learn the desired concept classes using back-propagation, while using the hypothesized control structure. Figure \ref{fig:plots} shows the activations of 10-d LSTM and GRU trained to recognize the languages $a^nb^n$ and $a^nb^nc^n$. It is clear that the LSTM learned to dedicate specific dimensions for counting, in contrast to the GRU.\footnote{Is the ability to perform unbounded counting relevant to ``real world'' NLP tasks? In some cases it might be. For example, processing linearized parse trees \cite{linp1,linp2,linp3} requires counting brackets and nesting levels. Indeed, previous works that process linearized parse trees report using LSTMs and not GRUs for this purpose. Our work here suggests that this may not be a coincidence.}

\section{The RNN Models}

An RNN is a parameterized function $R$ that takes as input an input vector $x_t$ and a state vector $h_{t-1}$ and returns a state vector $h_{t}$:
\begin{equation}
h_{t} = R(x_t, h_{t-1})
\label{eq:rnn}
\end{equation}

The RNN is applied to a sequence $x_1,...,x_n$ by starting with an initial vector $h_0$ (often the 0 vector) and applying R repeatedly according to equation (\ref{eq:rnn}).
Let $\Sigma$ be an input vocabulary (alphabet), and assume a mapping $E$ from every vocabulary item to a vector $x$ (achieved through a 1-hot encoding, an embedding layer, or some other means). Let $RNN(x_1,...,x_n)$ denote the state vector $h$ resulting from the application of $R$ to the sequence $E(x_1),...,E(x_n)$. An \emph{RNN recognizer} (or \emph{RNN acceptor}) has an additional function $f$ mapping states $h$ to ${0,1}$. Typically, $f$ is a log-linear classifier or multi-layer perceptron. We say that \emph{an RNN recognizes a language L$\subseteq \Sigma^*$} if $f(RNN(w))$ returns 1 for all and only words $w=x_1,...,x_n \in L$.

\paragraph{Elman-RNN (SRNN)} In the Elman-RNN \cite{elman1990finding}, also called the Simple RNN (SRNN), the function $R$ takes the form of an affine transform followed by a tanh nonlinearity:

\begin{equation}
h_{t} = \tanh(Wx_t + Uh_{t-1} + b)
\label{eq:srnn}
\end{equation}

Elman-RNNs are known to be at-least finite-state. Siegelmann \shortcite{siegelmann-fsa} proved that the tanh can be replaced by any other squashing function without sacrificing computational power.

\paragraph{IRNN} The IRNN model, explored by \cite{irnn}, replaces the $\tanh$ activation with a non-squashing ReLU: 
\begin{equation}
h_{t} = max(0,(Wx_t + Uh_{t-1} + b))
\label{eq:irnn}
\end{equation}
The computational power of such RNNs (given infinite precision) is explored in \cite{isi}.
\paragraph{Gated Recurrent Unit (GRU)} In the GRU \cite{cho2014learning}, the function $R$ incorporates a \emph{gating mechanism}, taking the form:
\begin{eqnarray}
z_{t} &=& \sigma(W^zx_t + U^zh_{t-1} + b^z) \\
r_{t} &=& \sigma(W^rx_t + U^rh_{t-1} + b^r) \\
\tilde{h}_t &=& \tanh(W^hx_t + U^h(r_t\circ h_{t-1}) +b^h) \\
h_{t} &=& z_t\circ h_{t-1}+(1-z_t)\circ\tilde{h}_t
\label{eq:gru}
\end{eqnarray}
\noindent Where $\sigma$ is the sigmoid function and $\circ$ is the Hadamard product (element-wise product).
\paragraph{Long Short Term Memory (LSTM)}
In the LSTM \cite{lstm}, $R$ uses a different gating component configuration:
\begin{eqnarray}
f_{t} &=& \sigma(W^fx_t + U^fh_{t-1} + b^f) \\
i_{t} &=& \sigma(W^ix_t + U^ih_{t-1} + b^i) \\
o_{t} &=& \sigma(W^ox_t + U^oh_{t-1} + b^o) \\
\tilde{c}_{t} &=& \tanh(W^cx_t + U^ch_{t-1} + b^c) \\
c_{t} &=& f_t \circ c_{t-1} + i_t\circ \tilde{c}_t \\
h_{t} &=& o_t\circ g(c_t)
\label{eq:lstm}
\end{eqnarray}
where $g$ can be either tanh or the identity.

\paragraph{Equivalences} The GRU and LSTM are at least as strong as the SRNN: by setting the gates of the GRU to $z_t=0$ and $r_t=1$ we obtain the SRNN computation. Similarly by setting the LSTM gates to $i_t=1$,$o_t=1$, and $f_t=0$. This is easily achieved by setting the matrices $W$ and $U$ to 0, and the biases $b$ to the (constant) desired gate values.

Thus, all the above RNNs can recognize finite-state languages.

\section{Power of Counting}
Power beyond finite state can be obtained by introducing counters. Counting
languages and k-counter machines are discussed in depth in \cite{fischer1968}.
When unbounded computation is allowed, a 2-counter machine has Turing power.
However, for computation bound by input length (real-time) there is a more interesting
hierarchy. In particular, real-time counting languages cut across the
traditional Chomsky hierarchy: real-time k-counter machines can recognize at least one
context-free language ($a^nb^n$), and at least one context-sensitive one ($a^nb^nc^n$).
However, they \emph{cannot} recognize the context free language given by the
grammar $S\rightarrow  x | aSa | bSb$ (palindromes).

\paragraph{SKCM} For our purposes, we consider a simplified variant of k-counter
machines (SKCM).
A counter is a device which can be incremented by a fixed amount (\textsc{Inc}),
decremented by a fixed amount (\textsc{Dec}) or compared to 0 (\textsc{Comp0}).
Informally,\footnote{Formal definition is given in Appendix A.} an SKCM is a finite-state automaton extended with $k$ counters, where
at each step of the computation each counter can be incremented, decremented or
ignored in an input-dependent way, and state-transitions and accept/reject decisions can inspect the
counters' states using \textsc{Comp0}.
The results for the three languages discussed above hold for the SKCM variant as well, with proofs provided in Appendix A.

\section{RNNs as SKCMs}
In what follows, we consider the effect on the state-update equations on a single dimension, $h_t[j]$. We omit the index $[j]$ for readability.

\paragraph{LSTM}
The LSTM acts as an SKCM by designating $k$ dimensions of the memory cell $c_t$ as counters. 
In non-counting steps, set $i_t=0, f_t=1$ through equations (8-9). In counting
steps, the counter direction (+1 or -1) is set in $\tilde{c_t}$ (equation 11)
based on the input $x_t$ and state $h_{t-1}$. The counting itself is performed
in equation (12), after setting $i_t=f_t=1$. The counter can be reset to 0 by
setting $i_t=f_t=0$.

Finally, the counter values are exposed through $h_t = o_tg(c_t)$, making it
trivial to compare the counter's value to 0.\footnote{Some further remarks on the LSTM: LSTM supports both increment and decrement in a single dimension. The counting dimensions in $c_t$ are exposed through a function $g$. For both $g(x)=x$ and $g(x)=\tanh(x)$, it is trivial to do compare 0. Another operation of interest is comparing two counters (for example, checking the difference between them). This cannot be reliably achieved with $g(x)=\tanh(x)$, due to the non-linearity and saturation properties of the $\tanh$ function, but is possible in the $g(x)=x$ case. LSTM can also easily set the value of a counter to 0 in one step. The ability to set the counter to 0 gives slightly more power for real-time recognition, as discussed by \citet{fischer1968}.

\textbf{Relation to known architectural variants}: Adding peephole connections \cite{peepholes} essentially sets $g(x)=x$ and allows comparing counters in a stable way. Coupling the input and the forget gates ($i_t = 1-f_t$) \cite{odyssey} removes the single-dimension unbounded counting ability, as discussed for the GRU.}

We note that this implementation of the SKCM operations is achieved by saturating the
activations to their boundaries, making it relatively easy to reach and maintain in practice.

\paragraph{SRNN} 
The finite-precision SRNN cannot designate unbounded counting dimensions.

The SRNN update equation is:
\[
h_t = \tanh (Wx + Uh_{t-1} + b)
\]
\[
h_t[i] = \tanh (\sum_{j=1}^{d_x}W_{ij}x[j] + \sum_{j=1}^{d_h}U_{ij}h_{t-1}[j] + b[i])
\]

By properly setting U and W, one can get certain dimensions of $h$ to update according to the value of $x$, by $h_t[i] = \tanh(h_{t-1}[i] + w_ix+b[i])$. 
However, this counting behavior is within a $\tanh$ activation. Theoretically,
this means unbounded counting cannot be achieved without infinite precision. Practically,
this makes the counting behavior inherently unstable, and bounded to a
relatively narrow region. While the network could adapt to set $w$ to be small
enough such that counting works for the needed range seen in training without
overflowing the $\tanh$, attempting to count to larger $n$ will quickly leave
this safe region and diverge.

\paragraph{IRNN}
Finite-precision IRNNs can perform unbounded counting conditioned on input
symbols.  This requires representing each counter as two dimensions, and
implementing \textsc{Inc} as incrementing one dimension, \textsc{Dec} as
incrementing the other, and \textsc{Comp0} as comparing their difference to 0.
Indeed, Appendix A in \cite{isi} provides concrete IRNNs for recognizing the languages $a^nb^n$ and $a^nb^nc^n$. This makes IBFP-RNN with ReLU activation more powerful than IBFP-RNN with a squashing activation.
Practically, ReLU-activated RNNs are known to be notoriously hard to train because of the exploding gradient problem.

\paragraph{GRU}
Finite-precision GRUs cannot implement unbounded counting on a given dimension.
The $\tanh$ in equation (6) combined with the interpolation (tying $z_t$ and
$1-z_t$) in equation (7) restricts the range of values in $h$ to between -1 and 1,
precluding unbounded counting with finite precision. Practically, the GRU can
learn to count up to some bound $m$ seen in training, but will not generalize
well beyond that.\footnote{One such mechanism could be to divide a given dimension by $k>1$ at each symbol encounter, by setting $z_t=1/k$ and $\tilde{h_t}=0$. Note that the inverse operation would not be implementable, and counting down would have to be realized with a second counter.} Moreover, simulating forms of counting behavior in equation (7) require
consistently setting the gates $z_t, \; r_t$ and the proposal $\tilde{h}_t$ to
precise, non-saturated values, making it much harder to find and maintain stable solutions.

\paragraph{Summary} We show that LSTM and IRNN can implement unbounded counting in dedicated counting dimensions, while the GRU and SRNN cannot. This makes the LSTM and IRNN at least as strong as SKCMs, and strictly stronger than the SRNN and the GRU.\footnote{One can argue that other counting mechanisms---involving several dimensions---are also possible. Intuitively, such mechanisms cannot be trained to perform unbounded counting based on a finite sample as the model has no means of generalizing the counting behavior to dimensions beyond those seen in training. We discuss this more in depth in Appendix B, where we also prove that an SRNN cannot represent a binary counter.}

\section{Experimental Results} \label{sec:exps}
Can the LSTM indeed learn to behave as a k-counter machine when trained using backpropagation?
We show empirically that:

\begin{enumerate}
\item LSTMs can be trained to recognize $a^nb^n$ and $a^nb^nc^n$.
\item These LSTMs generalize to much higher $n$ than seen in the training set (though not infinitely so).
\item The trained LSTM learn to use the per-dimension counting mechanism.
\item The GRU can also be trained to recognize $a^nb^n$ and $a^nb^nc^n$, but they do not have clear counting dimensions, and they generalize to much smaller $n$ than the LSTMs, often failing to generalize correctly even for $n$ within their training domain.
\item Trained LSTM networks outperform trained GRU networks on random test sets for the languages $a^nb^n$ and $a^nb^nc^n$.
\end{enumerate} 

Similar empirical observations regarding the ability of the LSTM to learn to recognize $a^nb^n$ and $a^nb^nc^n$ are described also in \cite{Gers2001}.

We train 10-dimension, 1-layer LSTM and GRU networks to recognize $a^nb^n$ and $a^nb^nc^n$. For $a^nb^n$ the training samples went up to $n=100$ and for $a^nb^nc^n$ up to $n=50$.\footnote{Implementation in DyNet, using the SGD Optimizer. Positive examples are generated by sampling $n$ in the desired range. For negative examples we sample 2 or 3 $n$ values independently, and ensuring at least one of them differs from the others. We dedicate a portion of the examples as the dev set, and train up to 100\% dev set accuracy.}

\paragraph{Results}
On $a^nb^n$, the LSTM generalizes well up to $n=256$, after which it accumulates a deviation making it reject $a^nb^n$ but recognize $a^nb^{n+1}$ for a while, until the deviation grows.\footnote{These fluctuations occur as the networks do not fully saturate their gates, meaning the LSTM implements an imperfect counter that accumulates small deviations during computation, e.g.: increasing the counting dimension by 0.99 but decreasing only by 0.98.  Despite this, we see that the its solution remains much more robust than that found by the GRU --- the LSTM has learned the essence of the counting based solution, but its implementation is imprecise.} 
The GRU does not capture the desired concept even within its training domain: accepting $a^nb^{n+1}$ for $n>38$, and also accepting $a^nb^{n+2}$ for $n>97$. It stops accepting $a^nb^n$ for $n>198$.

On $a^nb^nc^n$ the LSTM recognizes well until $n=100$. It then starts accepting also $a^nb^{n+1}c^n$. At $n>120$ it stops accepting $a^nb^nc^n$ and switches to accepting $a^nb^{n+1}c^n$, until at some point the deviation grows.
The GRU accepts already $a^9b^{10}c^{12}$, and stops accepting $a^nb^nc^n$ for $n>63$.

Figure \ref{fig:plots}a plots the activations of the 10 dimensions of the $a^nb^n$-LSTM for the input $a^{1000}b^{1000}$. While the LSTM misclassifies this example, the use of the counting mechanism is clear. Figure \ref{fig:plots}b plots the activation for the $a^nb^nc^n$ LSTM on $a^{100}b^{100}c^{100}$. Here, again, the two counting dimensions are clearly identified---indicating the LSTM learned the canonical 2-counter solution---although the slightly-imprecise counting also starts to show. In contrast, Figures \ref{fig:plots}c and \ref{fig:plots}d show the state values of the GRU-networks. The GRU behavior is much less interpretable than the LSTM. In the $a^nb^n$ case, some dimensions may be performing counting within a bounded range, but move to erratic behavior at around $t=1750$ (the network starts to misclassify on sequences much shorter than that). The $a^nb^nc^n$ state dynamics are even less interpretable.

Finally, we created 1000-sample test sets for each of the languages. For $a^nb^n$ we used words with the form $a^{n+i}b^{n+j}$ where $n\in\mathrm{rand}(0,200)$ and $i,j\in\mathrm{rand}(-2,2)$, and for $a^nb^nc^n$ we use words of the form $a^{n+i}b^{n+j}c^{n+k}$ where $n\in\mathrm{rand}(0,150)$ and $i,j,k\in\mathrm{rand}(-2,2)$.  The LSTM's accuracy was 100$\%$ and 98.6$\%$ on $a^nb^n$ and $a^nb^nc^n$ respectively, as opposed to the GRU's 87.0$\%$ and 86.9$\%$, also respectively.

All of this empirically supports our result, showing that IBFP-LSTMs can not only theoretically implement ``unbounded" counters, but also learn to do so in practice (although not perfectly), while IBFP-GRUs do not manage to learn proper counting behavior, even when allowing floating point computations.

\section{Conclusions}

We show that the IBFP-LSTM can model a real-time SKCM, both in theory and in practice. This makes it more powerful than the IBFP-SRNN and the IBFP-GRU, which cannot implement unbounded counting and are hence restricted to recognizing regular languages. The IBFP-IRNN can also perform input-dependent counting, and is thus more powerful than the IBFP-SRNN. 

We note that in addition to theoretical distinctions between architectures, it is important to consider also the practicality of different solutions: how easy it is for a given architecture to discover and maintain a stable behavior in practice. We leave further exploration of this question for future work.

\section*{Acknowledgments}
The research leading to the results presented in this paper is  supported by the European Union's Seventh Framework Programme (FP7) under grant agreement no. 615688 (PRIME), The Israeli Science Foundation (grant number 1555/15), and The Allen Institute for Artificial Intelligence.

\clearpage
\appendix

% \section{Supplementary Material}
% \label{sec:supplemental}
\section*{Appendix}
\section{Simplified K-Counter Machines}
We use a simplified variant of the k-counter machines (SKCM) defined in
\cite{fischer1968}, which has no autonomous states and makes classification decisions based on a combination of its current state and counter values. This variant consumes input sequences on a symbol by symbol basis, updating at each step its state and its counters, the latter of which may be manipulated by increment, decrement, zero, or no-ops alone, and observed only by checking equivalence to zero. To define the transitions of this model its accepting configurations, we will introduce the following notations:

\emph{Notations} We define $z:\mathbb{Z}^k\rightarrow\{0,1\}^k$ as follows: for every $n\in\mathbb{Z}^k$, for every $1\leq i\leq k$, $z(n)_i=0$ iff $n_i=0$ (this function masks a set of integers such that only their zero-ness is observed). For a vector of operations, $o\in\{-1,+1,\times 0,\times 1\}^k$, we denote by $o(n)$ the pointwise application of the operations to the vector $n\in\mathbb{Z}^k$, e.g. for $o=(+1,\times 0,\times 1)$, $o((5,2,3))=(6,0,3)$. 

We now define the model. An \emph{SKCM} is a tuple $M=\langle\Sigma,Q,q_o,k,\delta,u,F\rangle$ containing:

\begin{enumerate}
\setlength{\itemsep}{0pt}
\setlength{\parskip}{0pt}
\setlength{\parsep}{0pt}
\item A finite input alphabet $\Sigma$
\item A finite state set $Q$
\item An initial state $q_0\in Q$
\item $k\in \mathbb{N} $, the number of counters
\item A state transition function $$\delta:Q\times \Sigma\times \{0,1\}^k\rightarrow Q$$
\item A counter update function\footnote{
We note that in this definition, the counter update function depends only on the input symbol. In practice we see that the LSTM is not limited in this way, and can also update according to some state-input combinations --- as can be seen when it it is taught, for instance, the language $a^nba^n$
We do not explore this here however, leaving a more complete characterization of the learnable models to future work.} 
$$u:\Sigma \rightarrow \{-1,+1,\times 0,\times 1\}^k$$
\item A set of accepting masked\footnote{i.e., counters are observed only by zero-ness.} configurations $$F\subseteq Q\times\{0,1\}^k$$
\end{enumerate}

The set of \emph{configurations} of an SKCM is the set $C=Q\times \mathbb{Z}^k$, and the initial configuration is $c_0=(q_0,\bar{0})$ (i.e., the counters are initiated to zero). The transitions of an SKCM are as follows: given a configuration $c_t=(q,n)$ ($n\in\mathbb{Z}^k$) and input $w_t\in\Sigma$, the next configuration of the SKCM is $c_{t+1}=(\delta(q,w_t,z(n)),u(w_t)(n))$. 

The language recognized by a k-counter machine is the set of words $w$ for which the machine reaches an accepting configuration --- a configuration $c=(q,n)$ for which $(q,z(n))\in F$.

Note that while the counters can and are increased to various non-zero values, the transition function $\delta$ and the accept/reject classification of the configurations observe only their zero-ness.

\subsection{Computational Power of SKCMs}
We show that the SKCM model can recognize the context-free and context-sensitive languages $a^nb^n$ and $a^nb^nc^n$, but not the context free language of palindromes, meaning its computational power differs from the language classes defined in the Chomsky hierarchy. Similar proofs appear in \cite{fischer1968} for their variant of the k-counter machine.

\paragraph{$a^nb^n$:}
We define the following SKCM over the alphabet $\{a,b\}$:
\begin{enumerate}
\setlength{\itemsep}{0pt}
\setlength{\parskip}{0pt}
\setlength{\parsep}{0pt}
\item $Q=\{q_a,q_b,q_r\}$
\item $q_0=q_a$
\item $k=1$
\item $u(a)=+1,\;u(b)=-1$
\item for any $z\in\{0,1\}$:\\ 
	$\delta(q_a,a,z)=q_a, \;\;\;\;\;\delta(q_a,b,z)=q_b,\\
     \delta(q_b,a,z)=q_r,\;\;\;\;\;\delta(q_b,b,z)=q_b\\
     \delta(q_r,a,z)=q_r,\;\;\;\;\;\delta(q_r,b,z)=q_r$
\item $C=\{(q_b,0)\}$
\end{enumerate}
The state $q_r$ is a rejecting sink state, and the states $q_a$ and $q_b$ keep track of whether the sequence is currently in the ``$a$'' or ``$b$'' phase. If an $a$ is seen after moving to the $b$ phase, the machine moves to (and stays in) the rejecting state. The counter is increased on input $a$ and decreased on input $b$, and the machine accepts only sequences that reach the state $q_b$ with counter value zero, i.e., that have increased and decreased the counter an equal number of times, without switching from $b$ to $a$.
It follows easily that this machine recognizes exactly the language $a^nb^n$.

\paragraph{$a^nb^nc^n$:}
We define the following SKCM over the alphabet $\{a,b\}$. As its state transition function ignores the counter values, we use the shorthand $\delta(q,\sigma)$ for $\delta(q,\sigma,z)$, for all $z\in\{0,1\}^2$.
\begin{enumerate}
\setlength{\itemsep}{0pt}
\setlength{\parskip}{0pt}
\setlength{\parsep}{0pt}
\item $Q=\{q_a,q_b,q_c,q_r\}$
\item $q_0=q_a$
\item $k=2$
\item $u(a)=(+1,\emptyset),\\u(b)=(-1,+1),\\u(c)=(\emptyset,-1)$
\item for any $z\in\{0,1\}$:\\ 
	$\delta(q_a,a)=q_a,\;\delta(q_a,b)=q_b,\;\;\delta(q_a,c)=q_r,\\
     \delta(q_b,a)=q_r,\;\;\delta(q_b,b)=q_b,\;\;\delta(q_b,c)=q_c,\\
     \delta(q_c,a)=q_r,\;\;\delta(q_c,b)=q_r,\;\;\delta(q_c,c)=q_c,\\
     \delta(q_r,a)=q_r,\;\;\delta(q_r,b)=q_r,\;\;\delta(q_r,c)=q_r$
\item $C=\{(q_c,0,0)\}$
\end{enumerate}
By similar reasoning as that for $a^nb^n$, we see that this machine recognizes exactly the language $a^nb^nc^n$. We note that this construction can be extended to build an SKCM for any language of the sort $a_1^na_2^n...a_m^n$, using $k=m-1$ counters and $k+1$ states.

\paragraph{Palindromes:}
We prove that no SKCM can recognize the language of palindromes defined over the alphabet $\{a,b,x\}$ by the grammar $S\rightarrow x|aSa|bSb$. The intuition is that in order to correctly recognize this language in an one-way setting, one must be able to reach a unique configuration for every possible input sequence over $\{a,b\}$ (requiring an exponential number of reachable configurations), whereas for any SKCM, the number of reachable configurations is always polynomial in the input length.\footnote{This will hold even if the counter update function can rely on any state-input combination.}

Let $M$ be an SKCM with $k$ counters. As its counters are only manipulated by steps of 1 or resets, the maximum and minimum values that each counter can attain on any input $w\in\Sigma^*$ are $+|w|$ and $-|w|$, and in particular the total number of possible values a counter could reach at the end of input $w$ is $2|w|+1$. This means that the total number of possible configurations $M$ could reach on input of length $n$ is $c(n)=|Q|\cdot (2n+1)^k$.

$c(n)$ is polynomial in $n$, and so there exists a value $m$ for which the number of input sequences of length $m$ over $\{a,b\}$ --- $2^m$ --- is greater than $c(m)$. It follows by the pigeonhole principle that there exist two input sequences $w_1\neq w_2\in \{a,b\}^m$ for which $M$ reaches the same configuration. This means that for any suffix $w\in\Sigma^*$, and in particular for $w=x\cdot w_1^{-1}$ where $w_1^{-1}$ is the reverse of $w_1$, $M$ classifies $w_1\cdot w$ and $w_2\cdot w$ identically---despite the fact that $w_1\cdot x \cdot w_1^{-1}$ is in the language and $w_2\cdot x \cdot w_1^{-1}$ is not. This means that $M$ necessarily does not recognize this palindrome language, and ultimately that no such $M$ exists.

Note that this proof can be easily generalized to any palindrome grammar over $2$ or more characters, with or without a clear `midpoint' marker.

\section{Impossibility of Counting in Binary}
While we have seen that the SRNN and GRU cannot allocate individual counting dimensions, the question remains whether they can count using a more elaborate mechanism, perhaps over several dimensions. We show here that one such mechanism ---  a binary counter --- is not implementable in the SRNN.

For the purposes of this discussion, we first define a binary counter in an RNN.

\paragraph{Binary Interpretation} In an RNN with hidden state values in the range $(-1,1)$, the \emph{binary interpretation} of a sequence of dimensions $d_1,...,d_n$ of its hidden state is the binary number obtained by replacing each positive hidden value in the sequence with a `1' and each negative value with a `0'. For instance: the binary interpretation of the dimensions 3,0,1 in the hidden state vector $(0.5,-0.1,0.3,0.8)$ is 110, i.e., 6.

\paragraph{Binary Counting} We say that the dimensions $d_1,d_2,...,d_n$ in an RNN's hidden state implement a \emph{binary counter} in the RNN if, in every transition, their binary interpretation either increases, decreases, resets to 0, or doesn't change.\footnote{We note that the SKCMs presented here are more restricted in their relation between counter action and transition, but prefer here to give a general definition. Our proof will be relevant even within the restrictions.}

A similar pair of definitions can be made for state values in the range $(0,1)$.

We first note intuitively that an SRNN would not generalize binary counting to a counter with dimensions beyond those seen in training --- as it would have no reason to learn the `carry' behavior between the untrained dimensions. We prove further that we cannot reasonably implement such counters regardless.

We now present a proof sketch that a single-layer SRNN with hidden size $n\geq 3$ cannot implement an $n$-dimensional binary counter that will consistently increase on one of its input symbols. After this, we will prove that even with helper dimensions, we cannot implement a counter that will consistently increase on one input token and decrease on another --- as we might want in order to classify the language of all words $w$ for which $\#_a(w)=\#_b(w)$.\footnote{Of course a counter could also be `decreased' by incrementing a parallel, `negative' counter, and implementing compare-to-zero as a comparison between these two. As intuitively no RNN could generalize binary counting behavior to dimensions not used in training, this approach could quickly find both counters outside of their learned range even on a sequence where the difference between them is never larger than in training.}

\emph{Consistently Increasing Counter:}
The proof relies on the linearity of the affine transform $Wx+Uh+b$, and the fact that `carry' is a non-linear operation. We work with state values in the range $(-1,1)$, but the proof can easily be adapted to $(0,1)$ by rewriting $h$ as $h'+0.5$, where $h'=h-0.5$ is a vector with values in the range $(-0.5,0.5)$.

Suppose we have a single-layer SRNN with hidden size $n=3$, such that its entire hidden state represents a binary counter that increases every time it receives the input symbol $a$. We denote by $x_a$ the embedding of $a$, and assume w.l.o.g. that the hidden state dimensions are ordered from MSB to LSB, e.g. the hidden state vector $(1,1,-1)$ represents the number 110=6. 

Recall that the binary interpretation of the hidden state relies only on the signs of its values. We use $p$ and $n$ to denote `some' positive or negative value, respectively. Then the number 6 can be represented by any state vector $(p,p,n)$.

Recall also that the SRNN state transition is  $$h_{t} = \tanh(Wx_t + Uh_{t-1} + b)$$ and consider the state vectors $(-1,1,1)$ and $(1,-1,-1)$, which represent 3 and 4 respectively. Denoting $\tilde{b}=Wx_a + b$, we find that the constants $U$ and $\tilde{b}$ must satisfy:
\begin{eqnarray*}
tanh(U(-1,1,1)+\tilde{b})&=(p,n,n)\\
tanh(U(1,-1,-1)+\tilde{b})&=(p,n,p)\end{eqnarray*}
As tanh is sign-preserving, this simplifies to:
\begin{eqnarray*}
U(-1,1,1)=(p,n,n)-\tilde{b}\\
U(1,-1,-1)=(p,n,p)-\tilde{b}\end{eqnarray*}
Noting the linearity of matrix multiplication and that $(1,-1,-1)=-(-1,1,1)$, we obtain:
$$U(-1,1,1)=U(-(1,-1,-1))=-U(1,-1,-1)$$
$$(p,n,n)-\tilde{b} = \tilde{b} - (p,n,p)$$
i.e. for some assignment to each $p$ and $n$, $2\tilde{b}=(p,n,n)+(p,n,p)$, and in particular $\tilde{b}[1]<0$.

Similarly, for $(-1,-1,1)$ and $(1,1,-1)$, we obtain
\begin{eqnarray*}
U(-1,-1,1)=(n,p,n)-\tilde{b}\\
U(1,1,-1)=(p,p,p)-\tilde{b}\end{eqnarray*}
i.e. 
$$(n,p,n)-\tilde{b}=\tilde{b}-(p,p,p)$$ 
or $2\tilde{b} = (p,p,p)+(n,p,n)$, and in particular that $\tilde{b}[1]>0$, leading to a contradiction and proving that such an SRNN cannot exist. The argument trivially extends to $n>3$ (by padding from the MSB).

We note that this proof does not extend to the case where additional, non counting dimensions are added to the RNN --- at least not without further assumptions, such as the assumption that the counter behave correctly for \emph{all}  values of these dimensions, reachable and unreachable. One may argue then that, with enough dimensions, it could be possible to implement a consistently increasing binary counter on a \emph{subset} of the SRNN's state.\footnote{(By storing processing information on the additional, `helper' dimensions)} We now show a counting mechanism that cannot be implemented even with such `helper' dimensions.

\emph{Bi-Directional Counter:} We show that for $n\geq 3$, no SRNN can implement an $n$-dimensional binary counter that increases for one token, $\sigma_{up}$, and decreases for another, $\sigma_{down}$. As before, we show the proof explicitly for $n=3$, and note that it can be simply expanded to any $n>3$ by padding.

Assume by contradiction we have such an SRNN, with $m\geq 3$ dimensions, and assume w.l.o.g. that a counter is encoded along the first 3 of these. We use the shorthand $(v_1,v_2,v3)c$ to show the values of the counter dimensions explicitly while abstracting the remaining state dimensions, e.g. we write the hidden state $(-0.5,0.1,1,1,1)$ as $(-0.5,0.1,1)c$ where $c=(1,1)$. 

Let $x_{up}$ and $x_{down}$ be the embeddings of $\sigma_{up}$ and $\sigma_{down}$, and as before denote $b_{up}=Wx_{up}+b$ and $b_{down}=Wx_{down}+b$. Then for some reachable state $h_1\in\mathbb{R}$ where the counter value is $1$ (e.g., the state reached on the input sequence $\sigma_{up}$\footnote{(Or whichever appropriate sequence if the counter is not initiated to zero.)})), we find that the constants $U,b_{down}$, and $b_{up}$ must satisfy:
\begin{eqnarray*}
tanh(Uh_1+b_{up})=(n,p,n)c_1\\
tanh(Uh_1+b_{down})=(n,n,n)c_2
\end{eqnarray*}
(i.e., $\sigma_{up}$ increases the counter and updates the additional dimensions to the values $c_1$, while $\sigma_{down}$ decreases and updates to $c_2$.) Removing the sign-preserving function tanh we obtain the constraints 
\begin{eqnarray*}
Uh_1+b_{up}=(n,p,n)\mathrm{sign}(c_1)\\
Uh_1+b_{down}=(n,n,n)\mathrm{sign}(c_2)
\end{eqnarray*}
i.e. $(b_{up}-b_{down})[0:2]=(n,p,n)-(n,n,n)$, and in particular $(b_{up}-b_{down})[1]>0$.
Now consider a reachable state $h_3$ for which the counter value is $3$. Similarly to before, we now obtain
\begin{eqnarray*}
Uh_3+b_{up}=(p,n,n)\mathrm{sign}(c_3)\\
Uh_3+b_{down}=(n,p,n)\mathrm{sign}(c_4)
\end{eqnarray*}
from which we get $(b_{up}-b_{down})[0:2]=(p,n,n)-(n,p,n)$, and in particular $(b_{up}-b_{down})[1]<0$, a contradiction to the previous statement. Again we conclude that no such SRNN can exist.

\bibliography{naaclhlt2018}

\begin{thebibliography}{18}
\expandafter\ifx\csname natexlab\endcsname\relax\def\natexlab#1{#1}\fi

\bibitem[{Aharoni and Goldberg(2017)}]{linp3}
Roee Aharoni and Yoav Goldberg. 2017.
\newblock \href {http://aclweb.org/anthology/P17-2021} {Towards string-to-tree
  neural machine translation}.
\newblock In \emph{Proceedings of the 55th Annual Meeting of the Association
  for Computational Linguistics (Volume 2: Short Papers)}, pages 132--140,
  Vancouver, Canada. Association for Computational Linguistics.

\bibitem[{Chen et~al.(2017)Chen, Gilroy, Knight, and May}]{isi}
Yining Chen, Sorcha Gilroy, Kevin Knight, and Jonathan May. 2017.
\newblock \href {http://arxiv.org/abs/1711.05408} {Recurrent neural networks as
  weighted language recognizers}.
\newblock \emph{CoRR}, abs/1711.05408.

\bibitem[{{Cho} et~al.(2014){Cho}, {van Merrienboer}, {Gulcehre}, {Bahdanau},
  {Bougares}, {Schwenk}, and {Bengio}}]{cho2014learning}
Kyunghyun {Cho}, Bart {van Merrienboer}, Caglar {Gulcehre}, Dzmitry {Bahdanau},
  Fethi {Bougares}, Holger {Schwenk}, and Yoshua {Bengio}. 2014.
\newblock Learning {{Phrase Representations}} using {{RNN
  Encoder{\textendash}Decoder}} for {{Statistical Machine Translation}}.
\newblock In \emph{Proceedings of the 2014 {{Conference}} on {{Empirical
  Methods}} in {{Natural Language Processing}} ({{EMNLP}})}, pages 1724--1734,
  Doha, Qatar. {Association for Computational Linguistics}.

\bibitem[{Choe and Charniak(2016)}]{linp2}
Do~Kook Choe and Eugene Charniak. 2016.
\newblock \href {https://aclweb.org/anthology/D16-1257} {Parsing as language
  modeling}.
\newblock In \emph{Proceedings of the 2016 Conference on Empirical Methods in
  Natural Language Processing}, pages 2331--2336, Austin, Texas. Association
  for Computational Linguistics.

\bibitem[{{Chung} et~al.(2014){Chung}, {Gulcehre}, {Cho}, and
  {Bengio}}]{chung2014empirical}
Junyoung {Chung}, Caglar {Gulcehre}, KyungHyun {Cho}, and Yoshua {Bengio}.
  2014.
\newblock Empirical {{Evaluation}} of {{Gated Recurrent Neural Networks}} on
  {{Sequence Modeling}}.
\newblock \emph{arXiv:1412.3555 {[}cs]}.

\bibitem[{{Elman}(1990)}]{elman1990finding}
Jeffrey~L. {Elman}. 1990.
\newblock \href {https://doi.org/10.1207/s15516709cog1402_1} {Finding
  {{Structure}} in {{Time}}}.
\newblock \emph{Cognitive Science}, 14(2):179--211.

\bibitem[{Fischer et~al.(1968)Fischer, Meyer, and Rosenberg}]{fischer1968}
Patrick~C. Fischer, Albert~R. Meyer, and Arnold~L. Rosenberg. 1968.
\newblock \href {https://doi.org/10.1007/BF01694011} {Counter machines and
  counter languages}.
\newblock \emph{Mathematical systems theory}, 2(3):265--283.

\bibitem[{Gers and Schmidhuber(2001)}]{Gers2001}
F.~A. Gers and E.~Schmidhuber. 2001.
\newblock \href {https://doi.org/10.1109/72.963769} {Lstm recurrent networks
  learn simple context-free and context-sensitive languages}.
\newblock \emph{Transactions on Neural Networks}, 12(6):1333--1340.

\bibitem[{Gers and Schmidhuber(2000)}]{peepholes}
F.~A. Gers and J.~Schmidhuber. 2000.
\newblock \href {https://doi.org/10.1109/IJCNN.2000.861302} {Recurrent nets
  that time and count}.
\newblock In \emph{Proceedings of the IEEE-INNS-ENNS International Joint
  Conference on Neural Networks. IJCNN 2000. Neural Computing: New Challenges
  and Perspectives for the New Millennium}, volume~3, pages 189--194 vol.3.

\bibitem[{Greff et~al.(2017)Greff, Srivastava, Koutník, Steunebrink, and
  Schmidhuber}]{odyssey}
K.~Greff, R.~K. Srivastava, J.~Koutník, B.~R. Steunebrink, and J.~Schmidhuber.
  2017.
\newblock \href {https://doi.org/10.1109/TNNLS.2016.2582924} {Lstm: A search
  space odyssey}.
\newblock \emph{IEEE Transactions on Neural Networks and Learning Systems},
  28(10):2222--2232.

\bibitem[{{Hochreiter} and {Schmidhuber}(1997)}]{lstm}
Sepp {Hochreiter} and J{\"u}rgen {Schmidhuber}. 1997.
\newblock Long short-term memory.
\newblock \emph{Neural computation}, 9(8):1735--1780.

\bibitem[{Hubara et~al.(2016)Hubara, Courbariaux, Soudry, El-Yaniv, and
  Bengio}]{binarized}
Itay Hubara, Matthieu Courbariaux, Daniel Soudry, Ran El-Yaniv, and Yoshua
  Bengio. 2016.
\newblock \href
  {http://papers.nips.cc/paper/6573-binarized-neural-networks.pdf} {Binarized
  neural networks}.
\newblock In D.~D. Lee, M.~Sugiyama, U.~V. Luxburg, I.~Guyon, and R.~Garnett,
  editors, \emph{Advances in Neural Information Processing Systems 29}, pages
  4107--4115. Curran Associates, Inc.

\bibitem[{{Le} et~al.(2015){Le}, {Jaitly}, and {Hinton}}]{irnn}
Quoc~V. {Le}, Navdeep {Jaitly}, and Geoffrey~E. {Hinton}. 2015.
\newblock A {{Simple Way}} to {{Initialize Recurrent Networks}} of {{Rectified
  Linear Units}}.
\newblock \emph{arXiv:1504.00941 {[}cs]}.

\bibitem[{Siegelmann(1999)}]{siegelmann-book}
Hava Siegelmann. 1999.
\newblock \href {https://doi.org/10.1007/978-1-4612-0707-8} {\emph{Neural
  Networks and Analog Computation: Beyond the Turing Limit}}, 1 edition.
\newblock Birkh\"auser Basel.

\bibitem[{Siegelmann(1996)}]{siegelmann-fsa}
Hava~T. Siegelmann. 1996.
\newblock \href {https://doi.org/10.1111/j.1467-8640.1996.tb00277.x} {Recurrent
  neural networks and finite automata}.
\newblock \emph{Computational Intelligence}, 12:567--574.

\bibitem[{Siegelmann and Sontag(1992)}]{ss92}
Hava~T. Siegelmann and Eduardo~D. Sontag. 1992.
\newblock \href {https://doi.org/10.1145/130385.130432} {On the computational
  power of neural nets}.
\newblock In \emph{Proceedings of the Fifth Annual {ACM} Conference on
  Computational Learning Theory, {COLT} 1992, Pittsburgh, PA, USA, July 27-29,
  1992.}, pages 440--449.

\bibitem[{Siegelmann and Sontag(1994)}]{ss94}
Hava~T. Siegelmann and Eduardo~D. Sontag. 1994.
\newblock \href {https://doi.org/10.1016/0304-3975(94)90178-3} {Analog
  computation via neural networks}.
\newblock \emph{Theor. Comput. Sci.}, 131(2):331--360.

\bibitem[{Vinyals et~al.(2015)Vinyals, Kaiser, Koo, Petrov, Sutskever, and
  Hinton}]{linp1}
Oriol Vinyals, Lukasz Kaiser, Terry Koo, Slav Petrov, Ilya Sutskever, and
  Geoffrey Hinton. 2015.
\newblock \href {http://dl.acm.org/citation.cfm?id=2969442.2969550} {Grammar as
  a foreign language}.
\newblock In \emph{Proceedings of the 28th International Conference on Neural
  Information Processing Systems - Volume 2}, NIPS'15, pages 2773--2781,
  Cambridge, MA, USA. MIT Press.

\end{thebibliography}
\bibliographystyle{acl_natbib}

% \bibliography{naaclhlt2018}
% \bibliographystyle{acl_natbib}

\end{document}